\documentclass{article}

\PassOptionsToPackage{numbers, compress}{natbib}
 \usepackage[preprint]{neurips_2026}

\usepackage[utf8]{inputenc} 
\usepackage[T1]{fontenc}    
\usepackage[colorlinks]{hyperref}       
\usepackage{url}            
\usepackage{booktabs}       
\usepackage{amsfonts}       
\usepackage{amsmath}
\usepackage{nicefrac}       
\usepackage{microtype}      
\usepackage[dvipsnames]{xcolor}         
\usepackage{amsfonts}
\usepackage{enumitem}
\usepackage{graphicx}
\usepackage{wrapfig}
\usepackage{caption}

\newcommand{\R}{\mathbb{R}}

\newcommand{\myparagraph}[1]{\textbf{#1}}

\title{
Reconstruction vs. Prediction Pre-Training of Discrete Video Representations for Continuous Control
}

\author{%
  Ron Kosova \\
  University of Helsinki \thanks{Work done while interning at Aalto University.} \\
  \texttt{ron.kosova@helsinki.fi}
  \And
   Francesco Croce \\
   ELLIS Institute Finland, Aalto University \\
   \texttt{francesco.croce@aalto.fi}
}

\begin{document}

\maketitle

\begin{abstract}
    Self-supervised video encoders are commonly trained to either reconstruct pixels or predict latent features, yet it remains unclear how these objectives shape representations for downstream dynamics. We present a capacity-matched study isolating this choice: two identical video encoders with a shared FSQ bottleneck are trained under autoencoding and V-JEPA objectives on a continuous control environment. We introduce a simple asymmetric discretization scheme to apply the same bottleneck to the predictive model. The resulting representations exhibit a clear trade-off: the autoencoder preserves substantially more physical state information and achieves slightly better one-step latent prediction, while the predictive encoder accumulates less error under autoregressive rollout. Structural probes suggest that this advantage is associated with greater robustness to low-level visual perturbations and stronger dependence on temporal context. These results indicate that predictive and reconstructive objectives induce distinct representations of dynamics, trading state fidelity for properties that may benefit long-horizon prediction.
\end{abstract}

\section{Introduction}

    Pre-trained video encoders have emerged as viable substrates for world modeling \cite{assran2025vjepa2}, raising the question of how pre-training objectives shape representations for downstream dynamics. Predictive objectives such as JEPA are motivated partly by their tendency to discard unpredictable nuisance detail \cite{lecun2022apt}, while reconstruction objectives have historically been successful in model-based reinforcement learning \cite{ha2018worldmodels, hafner2020dreamer, hafner2024dreamerv3}.
    Recent work has examined this question through different downstream properties. \citet{yeom2026videoworld} probe inverse dynamics across pre-trained image and video models, finding that most of the benefit comes from temporal video pre-training, with predictive objectives providing a modest advantage. \citet{nilaksh2026reconstruction} find that reconstruction-based models achieve better pixel fidelity but worse planning performance. Both comparisons, however, vary pre-training and architecture scale alongside the objective.
    
    In this work, we aim to isolate the role of pre-training objective on downstream performance by matching architecture and pre-training data. 
   We train two identical ViViT Model 1 \cite{arnab2021vivit} time-causal encoders \cite{atanov2026videoflextok}, differing only in objective (reconstruction vs. latent prediction), on video from SAC agents \cite{haarnoja2018sac} in DMC Walker \cite{tassa2018dmc}.
   Moreover, we focus on discrete representations, i.e. relying on a finite codebook via FSQ quantization \cite{mentzer2023fsq}, as these can improve stability in long-horizon latent rollouts \cite{baek2025djepa}. 
   While FSQ has been used in autoencoders \cite{atanov2026videoflextok}, we design a novel and simple approach to integrate it in JEPA-style training.
   For evaluation, we focus on forward dynamics and long-horizon error accumulation, motivated by the latent ``dreaming'' used in model-based RL \cite{ha2018worldmodels, hafner2020dreamer}.
   Our experiments reveal: (1) reconstruction retains more decodable physical state and achieves slightly better single-step local dynamics; (2) prediction accumulates less compounding error during long-horizon rollouts; and (3) structural probes demonstrate that predictive representations are more invariant to visual perturbations, whereas reconstructive tubelets are less sensitive to temporal context.

\section{Methods}
    \myparagraph{Data \& inputs.} We use shared training data for the encoders and a separate held-out dataset for all downstream experiments, generated from the DeepMind Control Walker environment \cite{tassa2018dmc}. Specifically, we train a SAC agent \cite{haarnoja2018sac} for 500k total timesteps on privileged state observations using the default Stable-Baselines3 \cite{stable-baselines3} MLP policy.
    From four checkpoints spanning training progress, we generate 25k 100-frame trajectories for encoder training and 4,165 trajectories for downstream experiments. From each training trajectory, we sample 12 8-frame clips, yielding 300k clips split 90/10 into training and evaluation sets. The downstream probes construct their training and evaluation samples directly from the held-out trajectories using their respective sampling procedures (App.~\ref{appendix:exp}). Thus, each clip is represented as an input tensor \(x\in\R^{8\times 3\times 84\times 84}\).

    \myparagraph{Encoder architecture.} For the encoders we follow the ViViT Model 1 architecture, augmented with time-causal self-attention, similar to 
    VideoFlexTok \cite{atanov2026videoflextok}, and an FSQ \cite{mentzer2023fsq} discretization bottleneck with levels \(L = [9, 9, 9, 5, 5, 5]\) which gives a codebook of 91,125 codes. 
    This shared encoder architecture contains approximately 4.9M trainable parameters, and allows us to compare pre-training objectives under matched capacity.
    In all experiments, the encoders use non-overlapping tubelet embeddings \cite{arnab2021vivit} spanning 2 frames. Spatially, the input is patched into 49 non-overlapping patches of size \(12 \times 12\). After spatio-temporal flattening we end up with an input sequence of size \(\mathbb{R}^{196 \times 864}\) before encoding and \(\mathbb{R}^{196 \times 6}\) after.
    The FSQ discretization $\mathbf q : \R^6\rightarrow \R^6$ first applies a continuous bounded projection $f:\R^6\rightarrow \R^6$ for each per-dimension level $L_i$, then a rounding operation, i.e.
    \begin{align}
        f_i(z)=\left\lfloor L_i / 2 \right\rfloor\tanh(z_i), \quad \textrm{for} \; i=1, \ldots, 6, \quad \textrm{and} \quad \mathbf{q}(z)=\operatorname{round}(f(z)).
    \end{align}
    Alternative discretization approaches exist for both autoencoding \cite{oord2018vqvae} and JEPA-based models \cite{baek2025djepa}. However, we choose FSQ because it can be applied consistently across both paradigms while providing compact codes with regular geometry.

    \myparagraph{Training.} The reconstructive encoder is trained with a standard autoencoding (AE) objective, while the predictive encoder follows the original V-JEPA objective \cite{bardes2024revisitingfeaturepredictionlearning}. We use FSQ \cite{mentzer2023fsq} as a shared discrete bottleneck between the two settings. FSQ is directly applied between the encoder and decoder in the autoencoding model. In the predictive model, the student-teacher structure introduces an ambiguity regarding which representation should be discretized. To maintain a shared bottleneck while minimally perturbing the original V-JEPA objective, we discretize the student representation but only apply FSQ's bounded continuous transformation to the teacher representation. This exposes the student to the discrete bottleneck, while providing continuous targets for the predictive objective (see Fig.~\ref{fig:vjepaarch} in App.~\ref{appendix:arch} for an illustration of the architecture).
    To the best of our knowledge, this asymmetric application of FSQ to a JEPA-style objective has not been explored previously, and represents a simpler way to integrate discretization into the JEPA paradigm than what existing methods propose \cite{baek2025djepa}.
    We formalize the asymmetric bottleneck by defining the student and teacher representations as
    \begin{align}
        \mathbf{z}_\textrm{s} = \operatorname{round}((f\circ \mathbf{e})(x_\textrm{unmasked}))\oplus\mathbf{m},\qquad
        \mathbf{z}_\textrm{t} = (f\circ\overline{\mathbf{e}})(x)
    \end{align}
    where \(\oplus\,\mathbf{m}\) indicates inserting the learnable mask token \(\mathbf{m}\) in the output at the masked positions, \(\mathbf{e}\) and \(\overline{\mathbf{e}}\) denote the student and the EMA teacher encoders respectively. 
    The resulting reconstructive ($\mathcal{L}_{\mathrm{AE}}$) and predictive ($\mathcal{L}_{\mathrm{JEPA}}$) training objectives are
    \begin{align}
        \mathcal{L}_{\mathrm{AE}} = \|(\mathbf d\circ\mathbf q\circ\mathbf e)(x)-x\|_2, \qquad
        \mathcal{L}_{\mathrm{JEPA}} = \|\mathbf p[\mathbf z_s]_{\mathcal M} -\operatorname{sg}[\mathbf z_t]_{\mathcal M}\|_1,
    \end{align}
    where 
    \(\mathbf{d}\) is the decoder, \(\mathbf{p}\) is the predictor, and \(\operatorname{sg}\) denotes the stop-gradient. We use \(\mathcal{M}\) to imply that the loss is computed only between the masked positions. We also apply a variance-floor regularization term to the student representations 
    to encourage codebook utilization. In our experiments, both models achieve \(99\%\) codebook usage, demonstrating the effectiveness of FSQ discretization.
    
\section{Experiments}

    We evaluate informational content, local and long-horizon dynamics, and selected structural properties of the representations learned with both pre-training objectives
    (details 
    in App.~\ref{appendix:exp}).

    \subsection{Information \& Local Dynamics}
        The simulated environment provides access to the agent's 
        physical state and actions at each timestep, allowing us to probe the information retained by the representations along three complementary dimensions: physical state information, action predictability, and local representation dynamics.

        \myparagraph{State information.} We train a linear probe on the flattened (frozen) discrete representations of each clip to predict the 24-dimensional physical state of its final frame, comprising body orientations, velocities, and height. 
        The performance is measured by mean \(R^2\) across the 24 state dimensions on the held-out data. We find that the reconstructive encoder retains substantially more decodable physical state information, attaining an \(R^2\) score of \(0.78\), compared to JEPA's \(0.47\) (Table~\ref{tab:probing}).

        \myparagraph{Policy decoding.} We train 
        a linear probe to predict the 6-dimensional action taken by the agent at that step. This measures how much information about the behavior policy's action is linearly accessible from the representation. Again, AE outperforms JEPA, with an \(R^2\) of \(0.44\) vs. \(0.23\) (Table~\ref{tab:probing}). 

        \myparagraph{Local dynamics.} We train 10.6M-parameter action-conditioned Transformers for both encoders to predict the representation of the subsequent 8-frame clip, offset by one temporal group (2 frames), given the current representation and final-frame action. Performance is measured using mean \(L_1\) prediction error. Despite the difference in state information, 
        the local latent dynamics between the two encoders are comparable (AE achieves 0.19 transition \(L_1\) loss and JEPA 0.21, see Table~\ref{tab:probing}).
        
        \begin{figure*}
            \centering
        \begin{minipage}[t]{.5\columnwidth}
                \captionof{table}{\textbf{State and policy decoding, local dynamics.} AE better retains state and behavior information ($R^2$), but local dynamics ($L^1$) are comparable.}
                \centering
                \small
                \begin{tabular}{lccc}
                    \toprule
                     encoder &  \(R^2_\textrm{state}\; (\uparrow)\)     &     \(R^2_\textrm{action}\; (\uparrow)\)  & $L^1$ loss $(\downarrow)$
                     \\
                     \midrule
                     AE       & 0.78    & 0.44     & 0.19
                     \\
                     JEPA     & 0.47    & 0.23     & 0.21
                     \\
                     \bottomrule
                \end{tabular}
                \label{tab:probing}
        \end{minipage}
        \hfill
        \begin{minipage}[t]{.48\columnwidth}
                \captionof{table}{\textbf{Context dependence} per temporal group. JEPA representations are much more sensitive to preceding context than AE's.}
                \centering
                \small
                \begin{tabular}{lccc}
                    \toprule
                     encoder &  \(D^\text{ctx}_2 / D^\text{adj}_2\)     & \(D^\text{ctx}_3 / D^\text{adj}_3\)       & \(D^\text{ctx}_4 / D^\text{adj}_4\) 
                     \\
                     \midrule
                     AE       & 0.11      & 0.04      & 0.06
                     \\
                     JEPA     & 0.91      & 0.84      & 0.89
                     \\
                     \bottomrule
                \end{tabular}
                \label{tab:ctxdep}
        \end{minipage}
        \end{figure*}
        
        \begin{figure}
            \centering
            \vspace{-4mm}
            \includegraphics[width=0.78\linewidth]{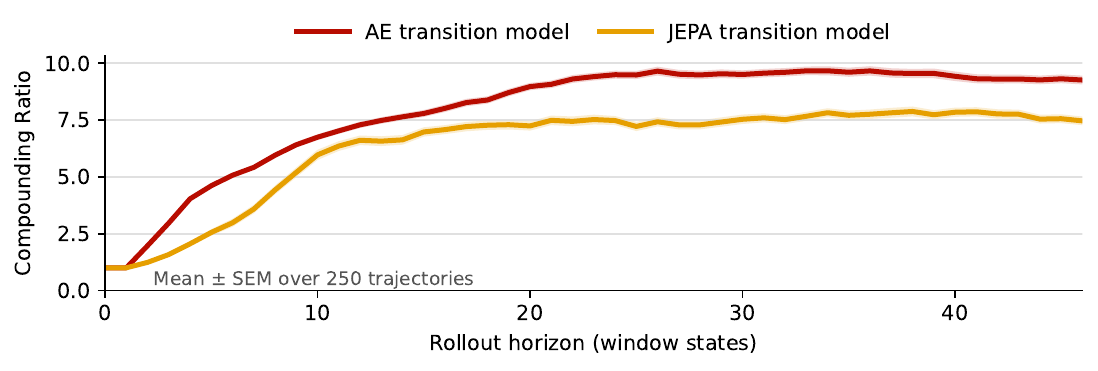} 
            \vspace{-4mm}
            \caption{\textbf{Compounding ratio of the transition models}, rolled out and averaged over 250 47-step trajectories: the transition trained on JEPA representations accumulates less error than the AE ones.}
            \label{fig:cr}
            \vspace{-3mm}
        \end{figure}

    \subsection{Long-Horizon Dynamics}
        A central motivation of JEPA-style objectives is that predictive learning encourages representations to discard low-level or otherwise unpredictable detail \cite{lecun2022apt, bardes2024revisitingfeaturepredictionlearning}. We hypothesize that this may also improve the error accumulation of autoregressive latent rollouts. We evaluate long-horizon error accumulation using the compounding ratio \cite{rao2026skyjepa}, defined as the ratio between the open-loop \(L_1\) error and the corresponding teacher-forced \(L_1\) error at rollout step \(k\). As shown in Fig.~\ref{fig:cr}, we find that, although the models exhibit comparable one-step transition error, the JEPA representation accumulates substantially less error over repeated transitions, achieving a mean CR of 6.5 over the trajectory, compared to the AE's 8.0. This is consistent with our hypothesis that the JEPA objective produces representations better suited to long-horizon autoregressive rollout. See App.~\ref{appendix:absdec} for absolute and decoded rollouts.

    \subsection{Structural Analysis}
    \label{sec:struct}
        \myparagraph{Context dependence.} We investigate whether the representation of a temporal segment depends on its preceding context. We independently encode two 8-frame windows offset by one temporal group (2 frames, as per tubelet depth size), yielding groups of spatial tokens \(g_1^t, \dots g_4^t\) and \(g_1^{t+1}, \dots, g_4^{t+1}\). Because the windows overlap, \(g_i^t\) and \(g_{i-1}^{t+1}\) encode the same 2-frame segment, but with different preceding temporal contexts. We measure the resulting context-induced representation shift as the mean \(L_1\) distance between corresponding spatial tokens ($D^\text{ctx}_i$). As a reference scale for this shift, we measure the mean distance between adjacent temporal groups within the same representation ($D^\text{adj}_i$):
         \begin{align}
            D^\text{ctx}_i &= \frac{1}{N}\sum_{j=1}^{N} \| g^t_i[j] - g^{t+1}_{i-1}[j] \|_1, &         D^\text{adj}_i &= \frac{1}{N}\sum_{j=1}^{N} \| g^t_i[j] - g^{t}_{i-1}[j] \|_1
        \end{align}
        where $N=49$ is the number of spatial positions. We report the ratios \(D_i^\mathrm{ctx}/D_i^\mathrm{adj}\), measuring context-induced variation relative to typical variation between adjacent temporal groups.
        For reference, a ratio value close to 1 indicates that the context-induced representation shift is comparable in magnitude to the typical shift between adjacent temporal groups, whereas a value close to 0 indicates that changing the preceding context induces comparatively little representation change relative to ordinary adjacent-group variation.
        Table~\ref{tab:ctxdep} indicates that JEPA representations (ratios $\approx 0.9$) are substantially more dependent on context compared to the AE representations (ratios $\approx 0.1$).
        This suggests that AE representations are largely determined by local visual content, whereas JEPA representations incorporate significantly more temporal context. Moreover, the ratios remain roughly constant across temporal groups ($i=2, 3, 4$) showing that the phenomenon is consistent across context lengths.

\begin{figure}
    \centering
    \includegraphics[width=0.8\linewidth]{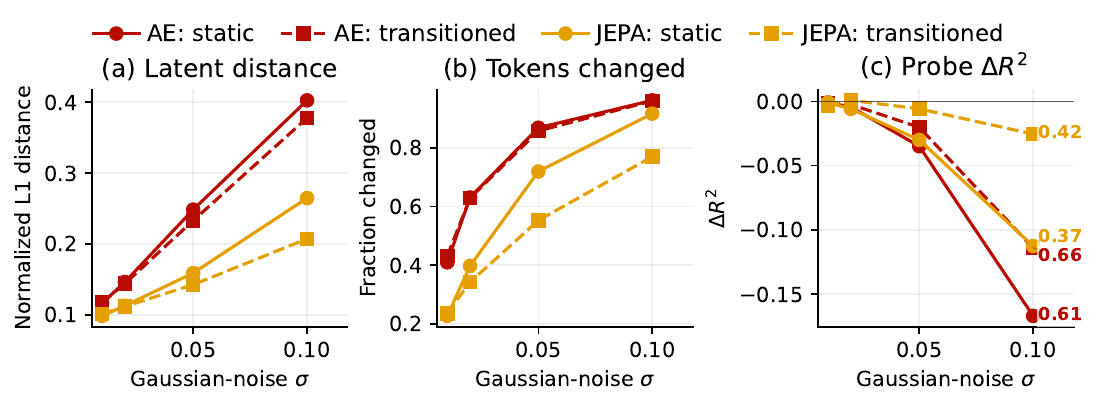}
    \vspace{-4mm}
    \caption{\textbf{Robustness to visual perturbations.} JEPA representations exhibit lower latent displacement, fewer changed tokens, and smaller state-probe degradation across perturbation strengths, both before and after a single transition.}
    \label{fig:perturbations}
    \vspace{-3mm}
\end{figure}

         \myparagraph{Robustness to visual perturbation.} Finally, we test the robustness of the representations to input perturbations.
         Then, we add zero-mean Gaussian noise of varying standard deviations $\sigma$ to the input frames, and measure the proportion of changed discrete tokens, the variance-normalized \(L_1\) distance between changed tokens, and degradation of the state probe. Measurements are performed both on static representations and after one transition. For transitioned representations, we reuse the local dynamics probe, rounding its outputs per dimension to the nearest integer to obtain the corresponding code.
         JEPA representations are more robust: 
         the average \(L_1\) distance of the changed tokens is lower both statically and through transition, fewer tokens change due to perturbation, and the state probe degradation is smaller (Fig.~\ref{fig:perturbations}). 
         Interestingly, the transition model itself appears to have a noise filtering effect, as the transitioned representations on perturbed inputs diverge from clean transitions less than static ones, especially for JEPA.
         We hypothesize that the higher robustness of JEPA transition model compared to AE one stems in the same structural properties of JEPA representation which yield its lower error accumulation (see Fig.~\ref{fig:cr}).

\section{Discussion \& Limitations}
    Our findings point toward a trade-off between visual and physical fidelity on one hand, and visual robustness and temporal context dependence on the other. This is consistent with the hypothesis that the abstraction induced by predictive objectives can benefit the modeling of long-term dynamics. Together with recent findings discussed above, our results provide further evidence that predictive objectives may be advantageous for representations intended for dynamical tasks in latent space, such as latent planning.
    These findings should nevertheless be interpreted in light of some limitations. Our models and datasets are comparatively small, and the study considers only a single continuous-control environment. 
    Moreover, we do not establish that the greater perturbation robustness or temporal context dependence of the predictive representation causes its improved long-horizon behavior; our structural probes instead identify properties that are consistent with such an explanation. Establishing the generality and causal role of these properties is an important direction for future work.

\bibliography{references}
\newpage
\appendix

\section{More on architectures, hyperparameters, and training}
\label{appendix:arch}

\begin{figure}
    \centering
    \includegraphics[width=1\linewidth]{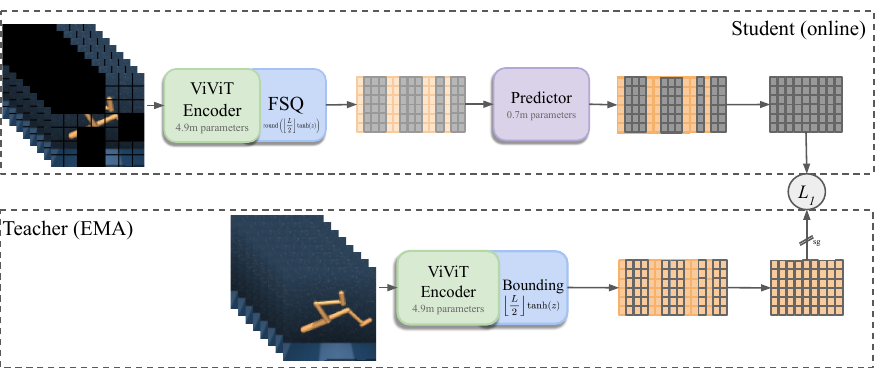}
    \caption{\textbf{Asymmetric FSQ bottleneck for V-JEPA}. The student representation is discretized with FSQ and masked before prediction, while the EMA teacher applies only the bounded continuous FSQ transformation, providing continuous prediction targets. The predictor is trained against the teacher representation at masked positions.}
    \label{fig:vjepaarch}
\end{figure}

In this appendix, we provide additional details on the architectures, hyperparameters, and training procedures of the two encoders used in our study.

\subsection{Encoder Architecture}

Both models share the same architecture: a ViViT Model 1 architecture, augmented with time-causal attention following VideoFlexTok \cite{atanov2026videoflextok}. Specifically, ViViT Model 1 extends the ViT architecture to video by embedding non-overlapping spatio-temporal tubelets and flattening the resulting tokens into a sequence \cite{arnab2021vivit}. For these embeddings, we use fixed 3D sinusoidal positional embeddings, constructed independently along the temporal and two spatial dimensions and concatenated to form each token's positional embedding.
These tokens then attend to each other in a time-causal manner, where each token in temporal group $t$ only attends to tokens in temporal groups $i\leq t$.
In both models, each token corresponds to a tubelet of shape $(2 \times 3\times 12\times 12)$, representing a spatial patch spanning two frames. The resulting tokens are passed through the shared FSQ bottleneck, which sits between the encoder and decoder in the AE and between the encoder and predictor in JEPA. In the case of the AE, the encoder and decoder are kept symmetrical, thus the decoder shares roughly the same number of parameters as the encoder (\(4.9\)M), whereas for JEPA we use a smaller, Transformer-based predictor (\(0.7\)M).

\subsection{Specifics of JEPA Masking}

\begin{figure}
    \centering
    \includegraphics[width=1\linewidth]{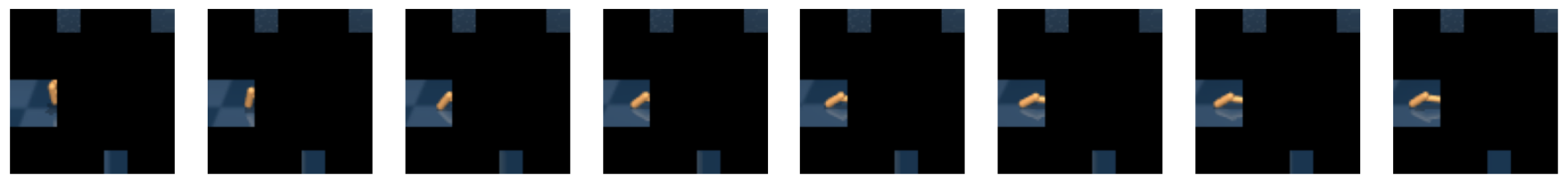}
    \includegraphics[width=1\linewidth]{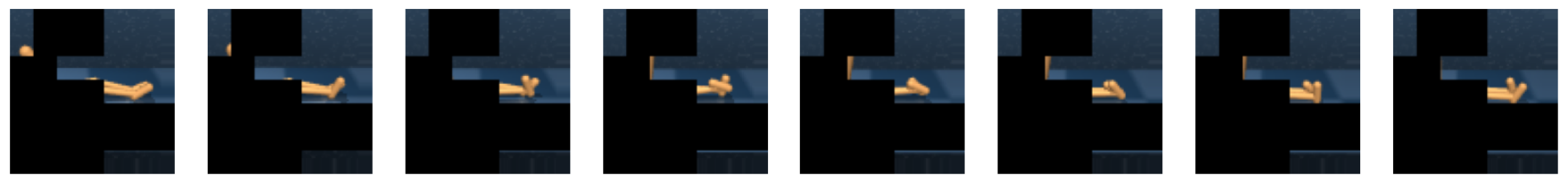}
    \caption{\textbf{Aggressive vs. relaxed masking.} The top example shows an instance of the masking strategy which covers \(\sim 90\%\) of the frame, whereas the bottom is an example of that which covers \(\sim 50\%\). The aggressive strategy proved unstable for the DMC Walker environment, where it is more likely to remove task-relevant context needed to predict the target blocks.}
    \label{fig:masks}
\end{figure}

The AE follows the standard autoencoding training procedure, while the JEPA model requires several additional design choices. We follow the training procedure described in \citet{bardes2024revisitingfeaturepredictionlearning}, with the addition of our asymmetric FSQ bottleneck. We also adopt the same masking strategy, sampling possibly overlapping contiguous spatial blocks of two types, "short-range" and "long-range." The union of these blocks is masked and thus used as the target, with its complement serving as context. However, we find the recipe used by \citet{bardes2024revisitingfeaturepredictionlearning} of \(8\) short-range masks and \(2\) long-range masks covering \(\sim 90\%\) of the frame unstable for our case. We hypothesize that this is because the masking is too aggressive for the DMC Walker environment: the walker occupies a relatively small portion of the frame, so aggressive masking can remove much of the task-relevant visual context. In this setting, the remaining context may provide insufficient information for prediction, potentially making collapse easier. We mitigate this in two ways. First, we relax the masking to 2 short-range and 2 long-range blocks, randomly sampled with relative sizes of \((0.1,0.25)\) and \((0.2,0.25)\), respectively, covering on average \(\sim50\%\) of the frame. Second, we randomly translate the input so that the Walker is not always centrally framed (see Fig.~\ref{fig:masks}). We find this sufficient to get the JEPA model to train without collapsing. 

\subsection{Training Details}

We strive to keep the training settings matched where possible, only deviating where necessitated by the innate differences of the two paradigms. See Tab.~\ref{tab:params}

\begin{table}[h]
    \caption{\textbf{Training and model hyperparameters.} Architectural and training settings for the autoencoding (AE) and predictive (JEPA) encoders used in our experiments.}
    \centering
    \label{tab:params}
    \vspace{2mm}
    \begin{tabular}{lcc}
        \toprule
        Parameter & AE & JEPA \\
        \midrule
        Encoder parameters             & \(4.9\)M & \(4.9\)M \\
        Decoder / predictor parameters  & \(4.9\)M & \(0.7\)M \\
        FSQ levels                      & \([9,9,9,5,5,5]\) & \([9,9,9,5,5,5]\) \\
        Input frames                    & \(8\) & \(8\) \\
        Tubelet size                    & \(2\) & \(2\) \\
        Spatial patch size              & \(12\times12\) & \(12\times12\) \\
        Optimizer                       & AdamW & AdamW \\
        Weight decay                    & \(0.05\) & \(0.05\) \\
        Batch size                      & \(350\) & \(400\) \\
        Warmup epochs                   & \(10\) & \(15\) \\
        Epochs                          & \(100\) & \(150\) \\
        Initial learning rate           & \(2\cdot10^{-5}\) & \(1.3\cdot10^{-5}\) \\
        Peak learning rate              & \(2\cdot10^{-4}\) & \(2\cdot10^{-4}\) \\
        LR warmup schedule              & Linear & Linear \\
        LR decay schedule               & Cosine & Cosine \\
        EMA initial momentum            & --- & \(0.996\) \\
        EMA final momentum              & --- & \(0.999\) \\
        EMA momentum schedule            & --- & Cosine \\
        Variance regularization weight  & --- & \(0.5\) \\
        \bottomrule
    \end{tabular}
\end{table}

The difference in batch size arises from the memory requirements of the respective model configurations and is therefore fixed by the available 12 GB GPU memory. We regard this as an inherent constraint of the compared configurations rather than an independently tuned hyperparameter. Epoch counts are chosen to approximately match total training time, with both models trained for approximately 9 hours on an RTX 3060 (12 GB).

\label{appendix:roll}
\begin{figure}
    \centering
    \vspace{-3mm}
    \includegraphics[width=1\linewidth]{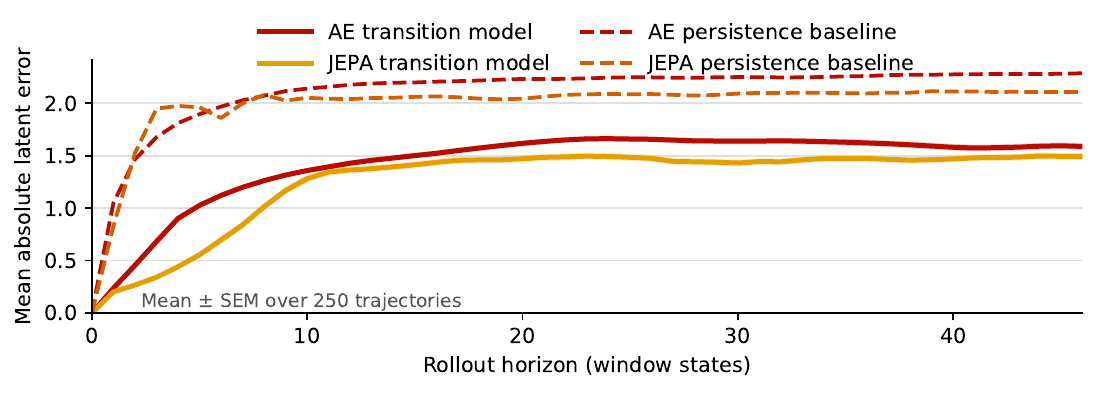}    
    \vspace{-3mm}
    \caption{\textbf{Rollout probe with absolute errors.} \textbf{Top:} Rollout probe, similar to Fig.~\ref{fig:cr}, using absolute error values and a no-op baseline.}
    \label{fig:abs}
\end{figure}

\section{Experimental Setup}
\label{appendix:exp}

Here we provide additional details on the settings of the data and probes used to conduct the experiments described above. 

\subsection{Downstream Dataset and Sampling}
We use the same unseen dataset for all downstream probes: a \(4,165\)-trajectory set sampled similarly to the encoder training data via trained SAC checkpoints. The probes sample from this dataset in different ways. For the state and policy decoding probes, we sample \(12\) clips per trajectory, yielding \(49,980\) clips, which we split 70/30 into training and validation sets. For the transition model, we sample one \(100\)-frame clip from each trajectory and divide each into \(8\)-frame windows with a stride of \(2\) frames. Each pair of subsequent windows forms a transition example, yielding approximately \(190\)k examples, again split 70/30. We use more training examples for the transition model due to the greater difficulty of the prediction task.

\subsection{State Decoding Probe Architecture}
We evaluate the physical state information contained in each representation using a linear probe trained on frozen encoder representations. The probe takes the flattened representation of an \(8\)-frame clip as input and predicts the \(24\)-dimensional physical state of the final frame, comprising body orientations, velocities, and height. The probe is trained on the training split of the downstream dataset, with performance evaluated on the held-out validation split. We report the mean \(R^2\) across the \(24\) state dimensions.

\subsection{Policy Decoding Probe Architecture}

We evaluate the extent to which the learned representations retain information relevant to the agent's policy using a policy decoding probe. The flattened representation of an \(8\)-frame clip is used to predict the \(6\)-dimensional action taken at its final frame. The probe is trained on the downstream training split and evaluated on the held-out split. We report the mean \(R^2\) across the six action dimensions.

\subsection{Transition Probe Architecture}
We use the same transition model for both encoders, consisting of a \(10.6\)M-parameter Transformer. Given the representation of an \(8\)-frame window and the \(6\)-dimensional action taken at its final frame, the model predicts the representation of the subsequent \(8\)-frame window, offset by one temporal group (\(2\) frames). The action is projected to the model's latent dimension and added to each input token before being processed by the Transformer. The model is trained using an \(L_1\) loss between predicted and target representations.

We use window-level transitions rather than predicting only the subsequent temporal group. This choice is motivated by the differing context dependence of the two representations (Section~\ref{sec:struct}). In particular, the lower context dependence of the AE representation means that restricting the transition model to the final temporal group could provide it with less temporal information than is available to the JEPA representation. Using the full \(8\)-frame window instead allows both representations to provide the temporal context they encode.

\begin{figure*}
    \centering
    \begin{minipage}[t]{1\textwidth}
        \centering
        \includegraphics[width=1\linewidth]{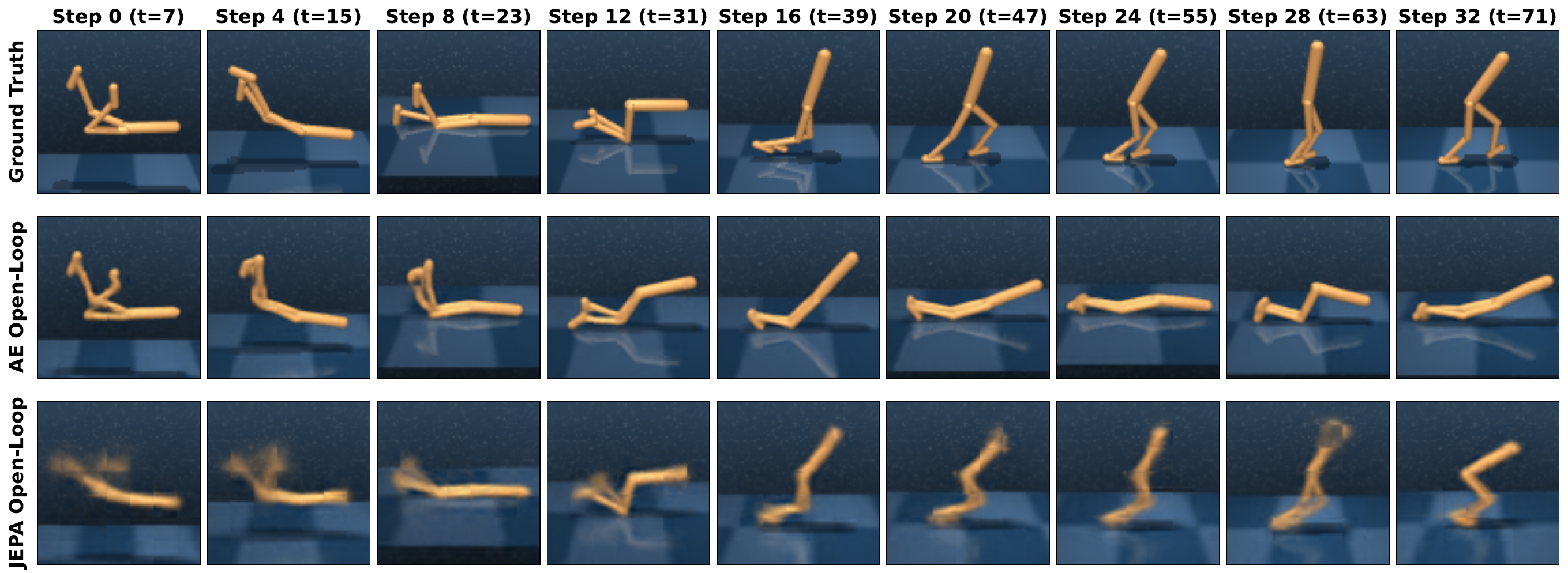}
    \end{minipage}
    \vfill
    \begin{minipage}[t]{0.32\textwidth}
        \centering
        \includegraphics[width=\linewidth]{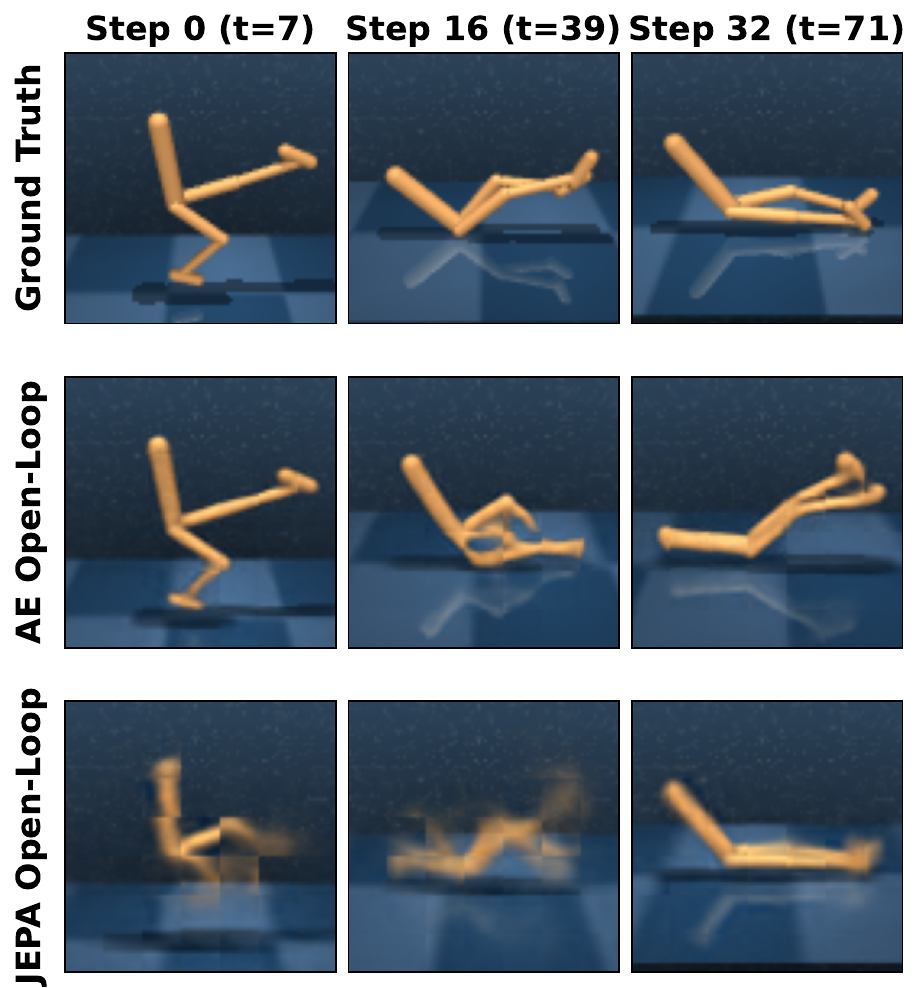}
    \end{minipage}
    \hfill
    \begin{minipage}[t]{0.32\textwidth}
        \centering
        \includegraphics[width=\linewidth]{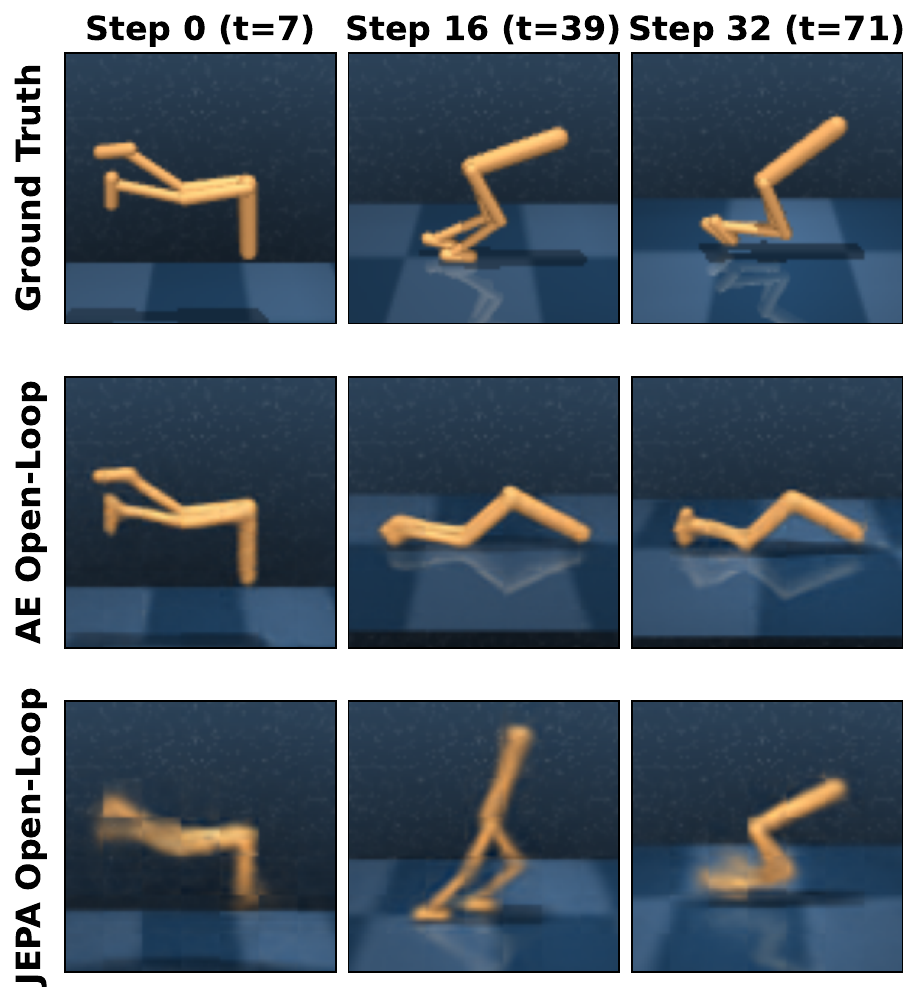}
    \end{minipage}
    \hfill
    \begin{minipage}[t]{0.32\textwidth}
        \centering
        \includegraphics[width=\linewidth]{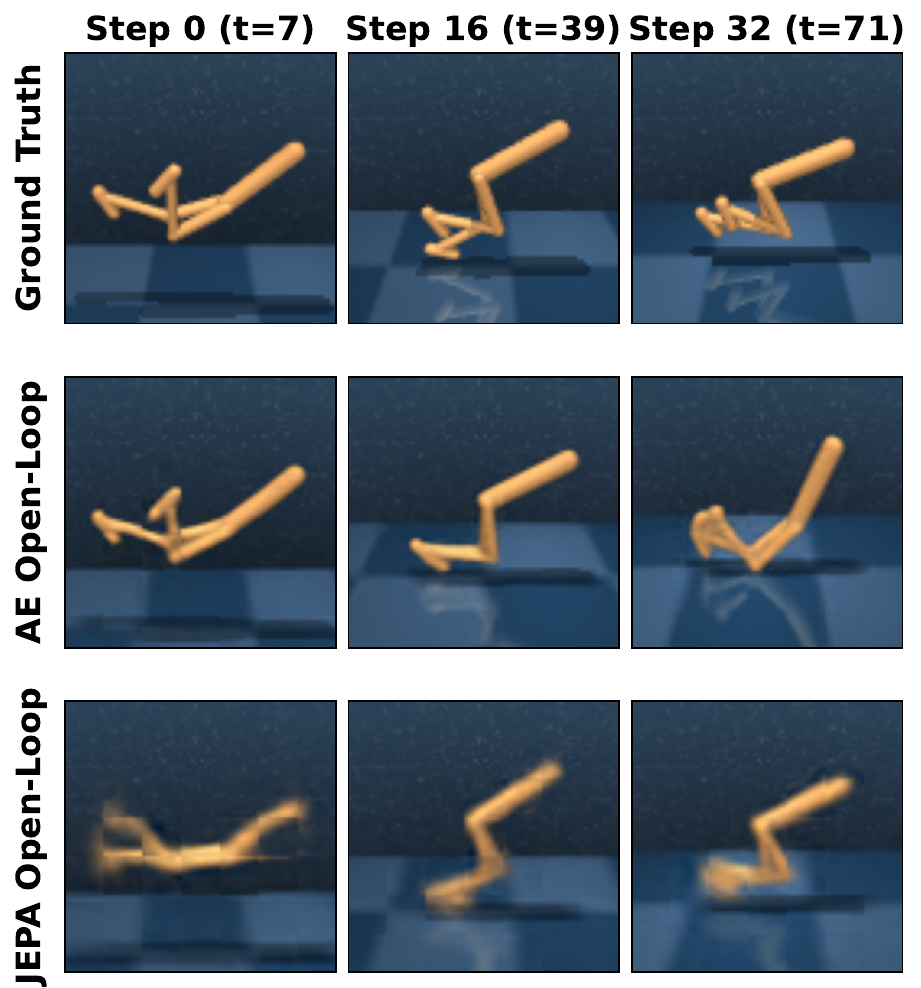}
    \end{minipage}
    \caption{\textbf{Decoded roll-outs.} Randomly selected trajectories rolled out and decoded with AE decoder (middle) vs. post-hoc JEPA decoder (bottom). All trajectories are of the same length. Note the visual consistency of the AE.}
    \label{fig:dec}
\end{figure*}

\section{Rollout Probe in Absolute Values and Decoded Rollouts}
\label{appendix:absdec}

Figure~\ref{fig:abs} represents the same data as the compounding ratio figure (Fig.~\ref{fig:cr}) in absolute error values. To account for differences between the two latent spaces, we also report a no-op baseline for both models (dashed lines). The baselines show that the absolute errors differ even without learned transition dynamics. With trained transitions (solid lines), the difference between the models becomes more pronounced up to a horizon of approximately 10 states, after which both models approach a similar plateau. 

We also provide samples of a qualitative analysis in Fig.~\ref{fig:dec} by decoding four randomly sampled trajectories using the pre-trained transition models. The autoencoder samples are decoded using the decoder which was trained jointly with the encoder, whereas for the JEPA model we post-train a decoder using the same architecture and data as the AE decoder for \(100\) epochs. The most striking difference between the two is that the autoencoder retains a lot more visual information which remains coherent throughout transitions. This is in line with the expected difference between the two objectives. In some of these specific trajectories, we also see that the JEPA model remains more accurate over the rollout. However, both of these analyses are confounded by factors such as, e.g., the differences between jointly trained and post-trained decoders.

\bibliographystyle{abbrvnat}
\end{document}